\documentclass[12pt, reqno, a4paper,oneside]{article}

\rmfamily{\fontsize{12pt}{\baselineskip}\selectfont}
\usepackage{multirow}
\usepackage{graphicx, epsfig, psfrag}
\usepackage{amsfonts,amsmath,amssymb,amsbsy,amsthm}
\usepackage{bm}
\usepackage{color}
\usepackage{float}
\usepackage{hyperref}
\usepackage{mathrsfs}
\usepackage{longtable}
\usepackage[top=3cm,bottom=3cm,left=3cm,right=3cm]{geometry}
\usepackage{epstopdf}
\usepackage{mathabx}
\usepackage{stmaryrd}
\usepackage{ulem}
\usepackage{scalerel}
\usepackage{enumitem}
\usepackage{algorithm}
\usepackage{subfigure}
\usepackage{booktabs}
\usepackage{threeparttable}
\usepackage{authblk}
\usepackage[utf8]{inputenc}

\usepackage{environments}
\numberwithin{theorem}{section}
\numberwithin{equation}{section}
\definecolor{jmcol}{rgb}{0, 0.7, 0}
\definecolor{mlcol}{rgb}{0.7, 0, 0}
\definecolor{todocol}{rgb}{0.0, 0.4, 0.0}

\newcommand{\tabincell}[2]{\begin{tabular}{@{}#1@{}}#2\end{tabular}} 

\title{Multi-objective optimization and explanation for stroke risk assessment in Shanxi province}
\author[a,b]{Jing Ma}
\author[c]{Yiyang Sun}
\author[d]{Junjie Liu}
\author[a,d,e]{Huaxiong Huang} 
\author[f]{Xiaoshuang Zhou}
\author[c]{Shixin Xu}
\affil[a]{Research Center for Mathematics, 
Beijing Normal University at Zhuhai, 519087, China}
\affil[b]{Laboratory of Mathematics and Complex Systems (Ministry of Education), 
School of Mathematical Sciences, 
Beijing Normal University, Beijing 100875, China}
\affil[c]{Duke Kunshan University, 8 Duke Ave, Kunshan, Jiangsu, China}
\affil[d]{BNU-UIC Joint Mathematical Research center,
Beijing Normal University, Zhuhai 519087, China}
\affil[e]{Department of Mathematics and Statistics, York University,
Toronto, Ontario, Canada} 
\affil[f]{Department of Nephrology, Shanxi Provincial People’s Hospital, Taiyuan, Shanxi, China;}
\date{}

\begin{document}
%\linenumbers
\maketitle

\begin{abstract}
%Chinese healthcare system has been significantly improved with the rapid economic development. However, China bears the heaviest stroke burden in the world.
Stroke is the top leading  causes of death in China (Zhou et al. The Lancet 2019).  
 A dataset from Shanxi Province is used 
to identify the risk of  each patient's at four states low/medium/high/attack and provide the state transition tendency through a SHAP DeepExplainer. 
%we gave the determinants of each patient's current risk state (low/medium/high/stroke) and provide the state transition tendency through a SHAP DeepExplainer based on a Deep Nerual Network (DNN) model. The DeepExplainer also produced the feature importance with respect to each risk state. % based on Multi-Layer Perception (MLP) as base model. 
To improve the accuracy on an imbalance sample set,
 the Quadratic Interactive Deep Neural Network (QIDNN) model  is first proposed by flexible selecting and appending of quadratic interactive features.
The experimental results showed that  the QIDNN model with 7 interactive features  achieve the state-of-art   accuracy  $83.25\%$. Blood pressure, physical inactivity, smoking, weight and total cholesterol are the top five important features.
Then, for the sake of  high recall on the most urgent state, attack state,  the stroke occurrence prediction is taken as an auxiliary objective to benefit from multi-objective optimization. The prediction accuracy was promoted, meanwhile the recall of the attack state was improved by $24.9\%$ (to $84.83\%$) compared to QIDNN (from $67.93\%$) with same features. The prediction model and analysis tool in this paper not only gave the theoretical optimized prediction method, but also provided the attribution explanation of risk states and transition direction of each patient, which provided a favorable tool for doctors to analyze and diagnose the disease.
\end{abstract}


%\todo{
%\begin{itemize}
%\item vocabulary: 
%	\begin{itemize}
%		\item order-2
%		\item stroke occurrence prediction and stroke risk prediction 
%		\item stroke state
%		\item current risk state and the state transition
%		\item low-risk
%		\item Expert1
%		\item transition
%	\end{itemize}
%\item citation: related work
%\item article: a/an and the
%\end{itemize}
%}

\section{Introduction}
\label{sec:intro}
Stroke is the second leading cause of death in people older than $60$ years and the fifth leading cause of death among those aged from 15 to 59 years old \cite{2011Stroke}. Over two-thirds of stroke deaths worldwide are in developing countries and almost one-third of which are in China \cite{liu2007stroke}. By 2017,  stroke is the top three   causes of death \cite{zhou2019mortality} and it accounted for 1.57 million deaths in 2018 \cite{2019survey}. The rising stoke patients also put huge pressure on the public health system.

Meanwhile, stroke is a preventable disease. A number of potent risk factors are  reported such as age, systolic blood pressure, smoking and so on \cite{lumley2002stroke}. Different algorithms are desired to increase the accuracy of risk prediction and effectiveness of disease prevention \cite{hung2017comparing}.
Traditional statistical methods, such as Framingham Stroke Risk Profile (FSRP) \cite{1991FSRP}, new FSRP \cite{2017FSRP}, the QStroke \cite{2013QStroke} and so on, aim to predict 10-year or 5-year risk of stroke. Most of these methods reflected the current state of stroke risk factors among the whites in the United States and France. In \cite{2019Stroke}, the prediction of 10-year or 5-year risk of stroke is evaluated in China. However, the performance heavily depends on the pre-selected features, and those methods are not able to effectively explore the complex non-linear relationships in data.

% This makes it possible that stroke risk can be predicted relatively accurately.
% A highly effective data-driven predictive algorithm is desired to increase the efficiency of disease prevention and improve patient outcomes through early detection and treatment \cite{hung2017comparing}.

Machine learning (ML) techniques are a set of powerful algorithms that are capable of modeling complex and hidden relationships between a multitude of clinical variables and the desired clinical outcome from data without stringent statistical assumptions. There is a growing interest in the application of machine learning techniques to address clinical problems \cite{fatima2017survey}. However, unlike the steady growth in the application of ML methods in other industries, the utilization of ML approach, especially for the use of deep learning in the  electronic health records (EHR) appears only recently.
%In stroke prediction, to the best of our knowledge, most researches are limited to stroke occurrence prediction using Logistic Regression (LR), Random Forest (RF), Support Vector Machine (SVM), decision trees and Deep Neural Networks (DNN) \cite{2021survey,hung2017comparing,khosla2010integrated,letham2015interpretable}. 
A high performance  DNN model \cite{2019largeEHR} is developed to predict 3 year and 8 year stroke occurrence    based on a large EHR dataset. It was also demonstrated in \cite{2019Machine} that DNN performed better than that of Logistic Regression (LR) and  Random Forest (RF), while predicted the long-term outcome for stroke occurrence.
Furthermore, combined with real-time electromyography bio-signals data,
\cite{2020RealTime} benefited from Long Short-Term Memory (LSTM) algorithm to balance the memory ratio between records (long-term) and real-time data (short-term). Most of  these methods are more focus on two states  classification, i.e. Stroke/Non-stroke. 

%From what mentioned above, deep learning performed better in stroke occurrence prediction, while there is few further work considering the states before stroke and few interpretation on cause of disease.
% \cite{2019largeEHR} obtain a high performance in predicting 3 year and 8 year stroke occurrence by combining the use of DNN model and a large EHR dataset.

%Compared with the methods above, deep learning is ver 
%difficult to comprehend due to the complexities in its framework. Furthermore, this method has not yet been demonstrated to achieve a better performance in small dataset comparing to other conventional ML algorithms in disease prediction tasks. 
Based on the census data in both communities and hospitals from Shanxi Province,  authors in \cite{series} proposed a model for three risk states (low/medium/high) prediction,  investigates different  stroke risk factors, including the ``8+2'' main risk factors proposed by the China National Stroke Prevention Project (CSPP),  and their ranking in Shanxi. It  shows  that  hypertension,  physical inactivity (lack of sports), and overweight are ranked as the top three high stroke risk factors in Shanxi.  

% on stroke prediction identified Chinese residents' risk levels of getting stroke to be low-risk, medium-risk and high-risk according to the ``8+2'' main risk factors proposed by the China National Stroke Prevention Project (CSPP), for the first time to the best of our knowledge. Meanwhile, with the help of Random Forest (RF) model, the ranking of influence factors in Shanxi Province were analyzed.

% In this paper,    a Quadratic Interactive Deep Neural Network (QIDNN) model for  the classification of four states "low/medium/high/stroke"  are proposed  for the first time to the best of our knowledge. Especially, SHAP  DeepExplainer  is developed to identify the transition probability 
% %\footnote{https://github.com/MadelineMa/MLCode/tree/master/Stroked.} The codes will be released if the paper is accepted. 
% The challenges of stroke prevention is the lack of accurate prediction and disease interpretability for each patient. The problem can be effectively solved through the model and analysis system provided in this paper.

 In this paper, we keep using the ``8+2'' main risk factors to identify the stroke risk state.
%  A Quadratic Interactive Deep Neural Network (QIDNN) model for  the classification of four states "low/medium/high/stroke"  are proposed  for the first time to the best of our knowledge.
Here the fourth state called attack (urgent state) is taken into consideration 
  with the data set of stroke attacked inpatients. As mentioned above, many studies considered the stroke occurrence, which motivate us to append the attack state. Moreover, we provide prediction and attribution on current risk state and state transition for each patient. For example, one patient is at a high-risk state, but the abnormal glycosylated hemoglobin and other factors make it possible to transfer to attack state. This requires us to append the attack state to those considered in \cite{series}, which are before the attack.
%based on the data, Figure \ref{fig:introstroke} shows the mean values of each state considered for the selected features, whose explaination are shown in Table \ref{tab:feature exp}. It shows that the stroke risk state have different representations with those states considered in \cite{series}, especially increasing the values of FBG and HbA1c to abnormal levels. These observations indicates that the stroke state could be considered as an individual state, moreover, the stroke state is more urgent that requires more careful consideration.}

%\begin{figure}
%\begin{center}
%	\includegraphics[width=14cm]{fig/intro_stroke_modi.png}
%	\caption{Mean values of the four stroke risk states for selected features}
%	\label{fig:introstroke}
%\end{center}
%\end{figure}

Our Main contributions are as follows:\\
1. A  DNN model, four-classification Multi-Layer Perceptrons (MLP),  is proposed  first as a baseline    to identify the current risk state.   Combined with the SHAP analysis tool, the determinants of each patient's current risk state and the state transition trend are given. \\
2. Then a Quadratic Interactive Deep Neural Network (QIDNN) model is established to impove the accuracy of the base DNN model by flexible selecting and appending of the quadratic interactive features for small sample set. Compared with the base DNN model, the mean of the recalls on four states increased from $74.48\%$ to $78.28\%$, the accuracy increased from $81.23\%$ to $83.25\%$ and the number of iterations decreased from 105 to 17 by adding 7 features to compute order-2 interactions to the QIDNN model.\\
3. Finally, a multi-objective model is  further refined to pay more attention to the most urgent state, attack state. The affects of sample unbalance was reduced alone with the better prediction of attack state.
%To ensure the convergence of the multi-objective model, we shrink the number of input by selecting Top 20 features ranking by feature importance. 
With the identical feature structure, the accuracy increased from $80.58\%$ to $84.45\%$ and the recall improved  from $67.93\%$ to $84.83\%$, compared to the QIDNN model with 3 order-2 related features.

This paper is organized as following. \S\ref{sec:framework} describes the proposed stroke risk prediction model and optimization methods, and \S\ref{sec:experiment} provides the experimental results. Finally, we conclude this work in \S\ref{sec:conc}.

\section{Materials and methods}
\label{sec:framework}

In this section, three models and a model interpretation method are introduced for the stroke risk prediction and intervention. 
We briefly review the material being used in the first place, and invoke the DNN model as the base model. Then, by considering the order-2 interactions, we propose the so-called QIDNN model to improve the performance for stroke risk prediction. Moreover, the stroke occurrence prediction is taken as an auxiliary objective to benefit from the MMOE model to achieve a higher recall. Finally, a model interpretation method called SHAP is introduced.

\subsection{Materials}
\label{subsec:dataset}

% China National Stroke Prevention Project (CSPP)
%To ensure high quality epidemiological studies, \cite{liu2007stroke} suggested a door-to-door survey. 
The dataset consists of  the survey data and  the
lab data of Shanxi province from CSPP, which is an original data set of 27,583 residents during the 2017 to 2020,  and 2,000 hospitalized stroke attacked patients in 2018. 
%All hospital records and survey data were reviewed by well-trained neurologists. 
After cleaning, 34 features are used, including gender, age and other basic features; smoking, lack of exercise (physical inactivity) and other life-style features; the lab data such as blood pressure. Table \ref{tab:feature exp} lists and explains some of the main features.
After cleaning, the numbers of  the four states (low:medium:high:attack) are $7,221: 5,868: 5,475:1,967$, respectively. If the low-, medium- and high-risk data are merged into the non-stroke category, the sample ratio of stroke and non-stroke is nearly 1:10.

\subsection{Base DNN model}
\label{subsec:basemodel}
The data with the labels of four states are used to train the model, which aims to predict the stroke risk of new patients.
\begin{figure}[!h]
	\subfigure[]{
		\label{fig:base_model}
		\includegraphics[width=7.5cm]{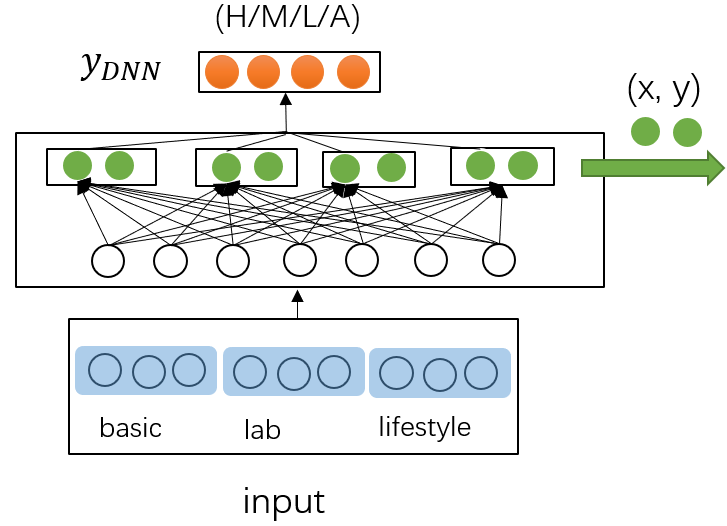}}
	\hspace{0.2cm} 
	\subfigure[]{
		\label{fig:overlap}
		\includegraphics[width=7.5cm]{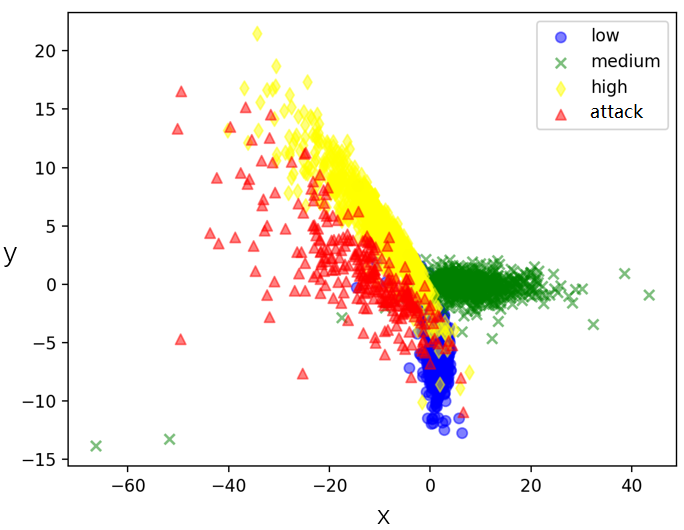}}
	\hspace{0.2cm}  
	\caption{Illustration of base DNN model. (a) Model structure. (b) Visualization of four classes.}
\end{figure}

The base DNN model of this paper adopts the multi classification model based on MLP of supervised learning, as shown in Figure \ref{fig:base_model}. The input contains 34 categorical-continuous-mixed original features. Due to the limited number of samples, the features are directly sent to the hidden layer composed of 17 neurons without the vectorization of one-hot or embedding. Finally, the output layer gives the predicted  probabilities of the four states through the {\it softmax} function.
\begin{displaymath}
	y_{DNN}=softmax(W^oa^h+b^o), 
\end{displaymath}
where $W^o$, $b^o$ are the model weight and bias of the output layer, $a^h$ is the output of the hidden layer. 

In order to give a better sense, before output, a hidden layer composed of $2\times 4$ neurons (makred green in Figure \ref{fig:base_model}) is appended for visualization. Two neurons, corresponding to one specific risk state, produces a location of the risk state in the two dimensional space.
Figure \ref{fig:overlap} shows the distribution of the four stroke risk states in the reduced dimensions. There are overlaps among these states as well as relevances and differences, but no inclusion relation.

\subsection{QIDNN}
\label{subsec:qidnn}
%By adding order-2 interactions of features, we aim to improve the model's ability of learning sophisticated feature interactions. At the same time, we need to control the number of parameters to ensure the convergence on the small sample set. If we use DeepFM directly, the number of parameters is the product of numbers of parameters contains in each layer, by embedding and the neural combination of FM and Deep components, the model may fail to converge due to a large amount of parameters. Moreover, DeepFM is more suitable for binary classification problem, there is little extended research on multi-classification problem.
\begin{figure}[!h]
	\begin{center}
		\includegraphics[width=14cm]{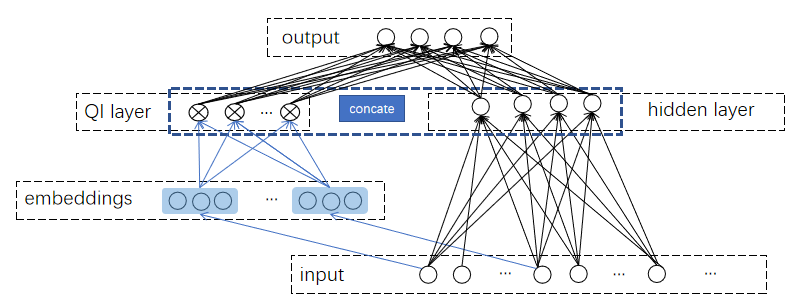}
		\caption{QIDNN model. }
		\label{figs:QIDNN}
	\end{center}
\end{figure}
In order to further improve the model's accuracy and ability of learning sophisticated feature interactions, 
a multi-classification model, called quadratic interactive deep neural network (QIDNN),   is proposed to predict the stroke risk with small dataset. The structure of QIDNN is shown in Figure \ref{figs:QIDNN}. It consists of two components, QI component and deep component. The original features enter the hidden layer for information extraction, and some latent vector $V_i$ is obtained from the selected features through the embedding layer. It is fed into the QI component to model the order-2 interactions to measure its impact with other features, and the output probability is produced again by the {\it softmax} function with the concatenation of both components as the input,

\begin{equation}
	y=softmax(y_{QI} \oplus y_{DNN}),\nonumber
	\label{eqn:yqidnn}
\end{equation}
where $y$ is the predicted value, $y_{QI}$ is the output of QI component, and $y_{DNN}$ is the output of deep component, and $\oplus$ means concatenate. The structure of the deep component is consistent with the base DNN model. 

QI component models pairwise feature interactions as inner product of respective feature latent vectors. It can capture order-2 feature interactions much more effectively. This idea is originated from DeepFM \cite{2017DeepFM}, while direct usage of DeepFM may fail to converge due to a large amount of parameters in our small dataset. As shown in Figure \ref{figs:QIDNN}, similar with Factorization Machines (FM) \cite{fm2010}, the output of QI component is a number of inner product units:
\begin{displaymath}
	y_{QI}=\sum_{i=1}^{n}\sum_{j=i+1}^{n}<V_i,V_j>x_{i}{x_{j}}\\
	=\frac{1}{2}\sum_{l=1}^{k}\left[\left(\sum_{i=1}^{n}v_{i,l}x_i\right)^2-\sum_{i=1}^{n}v_{i,l}^2x_i^2\right],
\end{displaymath}
where $n$ is the number of combination features in QI layer, $k$ is the length of latent vector, and $V_i$ is the latent vector representation of feature $x_i$, $v_{i,l}$ is the value of feature $i$ at the $l$-th position of the latent vector.

The main advantage of QIDNN is that by adding order-2 interactions into multi classification model, it could  flexibly controlling the number of combination features, balancing model parameters and sample number to ensure the model convergence in small dataset.
\subsection{MMOE}
\label{subsec:mmoe}

\begin{figure}[!h]
	\centering
	\includegraphics[width=9cm]{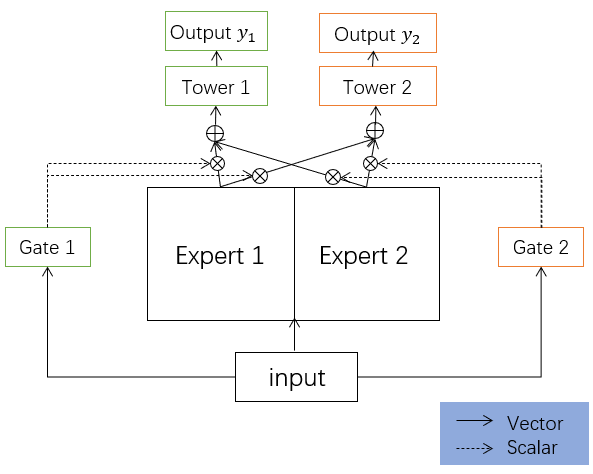}
	\caption{Multi-gate Mixture-of-Experts model. }
	\label{fig:MMOE}
\end{figure}

The recall of attack state is more important for clinical diagnosis, since every urgent case should be predict correctly. The MMOE model \cite{2018MMOE},  one of the most popular method of multi-objective learning, is used to further refine the attack state without increasing the number of model parameters obviously for the small dataset of our project.

%however, the precision and recall of the base models for this task are not good enough (see \S\ref{subsec:pred_rt}). Most of the existing researches considered the stroke occurrence prediction through binary classification \cite{khosla2010integrated,hung2017comparing}, however, they paid more attention to AUC \cite{ROC} and ignore the effect of recall. 

As shown in Figure \ref{fig:MMOE}, we apply the MMOE model to optimize the predictions of both stroke occurrence and stroke risk, simultaneously. The intermediate layers of DNN and QIDNN are embedded as Expert1 and Expert2, respectively. Combining with weights produced by gates, we input the weighted sum of comprehensive opinions of experts to towers to learn the characteristics of each objective. Tower1 returns the stroke occurrence prediction, and Tower2 outputs the stroke risk prediction. We note that, the stroke occurrence prediction is considered as an auxiliary objective to attract more attention on the attack state \cite{2017overviewMOOP}.

By MMOE, both objectives can be  through common information and specific information in between.
Concretely, in Figure \ref{fig:MMOE}, for objective $k\in\{1,2\}$, $y_k$ is the output of tower $k$, which consists of multi-layer full connections,
\begin{displaymath}
	y_k=h^k(f^k(x)),
\end{displaymath}
where $h^k$ is the $k$-th tower network. The input of the tower, $f^k(x)=\sum_{i=1}^{n}g^k_i(x)f_i(x)$, is the weighted sum of each expert by corresponding gate.  It indicates that the gating networks of the $k$-th objective realizes the selective utilization of experts by different weights. The $n$ expert networks are denoted as $\{f_i\}_{i=1}^{n}$, $g_i$ is the weight of $i_{th}$ expert on the final decision satisfies $\sum_{i=1}^{n}g_i(x)=1$. In the optimization of different objectives, the effects of different experts' opinions are adjusted through the gates.

%Therefore, different gating networks of different objectives can learn different patterns of combined experts, that is, the model takes into account the relevance and difference of captured objectives.

%Stroke occurrence prediction tells whether the patient will have stroke or not according to the binary classification model. By optimizing both the stroke occurrence prediction and stroke risk prediction simultaneously, the stroke state is emphasized in both objectives. 
%Then by learning similarity and difference between the objectives, the appropriate multi-objetive optimization could improved the results of both objectives. As a result, the prediction on stroke state was improved. Since the Pearson correlation between stroke occurrence prediction and stroke risk prediction is about 0.2, their correlation is relatively weak. In this paper we choose MMOE framework instead of shared bottom.

%the proportion of different experts opinions is achieved through the gates. 
Obviously, the MMOE model is more complex than any of its expert networks, extra parameters requires additional samples to converge. To cope with the small dataset, we take two measures to guarantee the convergence of MMOE. Firstly, we remove the towers, and feed the result directly to the output layer. The loss function is as following:
\begin{equation}
	L(\theta_1,\theta_2)=\sum_{i=1}^Nl_1(y_i,f(x_i;\theta_1))+\sum_{i=1}^Nl_2(z_i,f(x_i;\theta_2)),
\end{equation}
where $\theta_1$ and $\theta_2$ are the parameters of the stroke occurrence prediction and the stroke risk prediction, $l_1$ is the binary cross entropy function, $l_2$ is the cross entropy function, $y$ and $z$ are the labels of the stroke occurrence prediction and the stroke risk prediction, respectively.

%if the 34 features, used by the base DNN model, were input to the MMOE model, the existing sample size can not guarantee the convergence of the model, and the experiment shows that the loss value fluctuates dramatically.
Secondly, we rank the features according to their importances and select the top 20 features as the input. For the construction of the model, Expert1 contains only one single hidden layer with 11 neurons, and Expert2 adopts QIDNN model with 3 order-2 related features. The stroke occurrence prediction with one neural as Output $y_1$ and the stroke risk prediction with four neurons as Output $y_2$. The early stop mechanism is used to prevent over-fitting.

\subsection{SHAP}
\label{subsec:shap}
Feature importance refers to techniques that assign a score to input features based on how useful they are at predicting a target variable.
Traditional feature importance methods can only rank the importance of different features, without explaining how the feature effects the predictions ((positively or negatively). 
Shapley value is a method originated from cooperative game theory proposed by Shapley in 1953 \cite{shap1953definition}.
It is the only attribution method that satisfies the four attributes of efficiency, symmetry, dummy and additivity, which can be regarded as the definition of fair expenditure.

SHAP \cite{NIPS2017} is a ``model interpretation" package developed by Python, which can interpret the output of most machine learning models. Inspired by Shapley, SHAP constructs an additive interpretation model, which interprets the predictive value of the model as the sum of the attribution values of each input feature. All features are regarded as ``contributors", and SHAP value is the average marginal contribution of the features to the output of the model. 

Denote the $i$-th sample as $\boldsymbol{x}_i$ and the $j$-th feature of the $i$-th sample as $x_{ij}$. Let $y_i$ be the predicted value of the model for this sample, and let $y_{base}$ represent the baseline output of the model, then the following equation holds,
\begin{displaymath}
	y_i = y_{base}+f(x_{i1})+f(x_{i2})+...+f(x_{ik}),
\end{displaymath}
where $f(x_{ij})$ is the SHAP value of $x_{ij}$. Intuitively, $f(x_{i1})$ is the contribution of the first feature in the $i$-th sample to the final predicted value $y_i$. The value $f(x_{i1})>0$ indicates that the feature improves the predicted value with a positive effect. On the contrary, negative value of $f(x_{i1})$ tells that the feature has a negative effect that reduces the predicted value.

\section{Result}
\label{sec:experiment}
%In this section, a base DNN model is proposed for the risk (low/medium/high/attack)  assessment with a dataset from Shanxi Province. Then by adding order-2 interactions of features, a high accuracy QIDNN model is developed. In order to further improve the recall of attack state, MMOE  model is established.  Further more, we also provide a SHAP  DeepExplainer  based  on  the  QIDNN  model  to identify the dominate risk factors leading to state transition for each subject. 

\subsection{DNN}
 The performance of a based DNN model is  shown first for the stroke risk assessment.  Then, the features' importance for each state are  given by SHAP values.
\label{subsec:pred_rt}
\subsubsection{Numerical results of DNN}
\label{sssec:DNN}
At present, most interpretable models of machine leaning in healthcare are based on tree models \cite{ahmad2018interpretable,2017Consistent}, which promotes the application of tree model. In \cite{series} of our series work, random forest (RF) was invoked. As shown in Table \ref{tab:base}, RF works pretty well for first three states (low/medium/high-risk). However, the imbalance of data leads to poor recall of the attack state with the tree models. In Table \ref{tab:base},
the head ``support'' means the sample number of test set, where the support of attack state is less than the others. 
The explanations of the evaluation indicators are:

\begin{itemize}
    \item {\bf Accuracy}: the proportion of the number of correctly predicted samples over the total number of samples. It is the most intuitive indicator to measure the quality of the model.
    
	\item {\bf Precision}: the proportion of the number of positive samples correctly predicted over the number of positive samples predicted. The higher the precision, the more accurate the prediction.
	
	\item {\bf Recall}: the proportion of the number of the positive samples correctly predicted over the number of all positive labels. The higher the recall, the more complete the result.
	
	\item {\bf F1 score} $:=2\frac{\textrm{precision}\times \textrm{recall}}{\textrm{precision}+\textrm{recall}}$, is the harmonic mean of the precision and recall. 
\end{itemize}

From Table \ref{tab:base}, the DNN model can predict the attack state well, while the results of low-, medium- and high-risks are slightly weaker than the RF model. However, we still select the DNN model as the base model. In clinical diagnosis, the prediction of the attack state counts more, which means to correctly predict every sample being in the attack state. Hence, the DNN model, with a much higher recall of the attack state plays the role of the base model.

% did not select the RF method as the base model, for the relatively smaller recall, which is caused by the small amount of the imbalance samples. Although the precisions and recalls of the low/medium/high-risk states are better than the base DNN model, which being introduced in \S\ref{subsec:basemodel}, the prediction of stroke state counts more in clinical diagnosis, and we wish to correctly predict every sample being in stroke state, which requires a higher recall (see Table \ref{tab:base} for details).
%So, we set the base DNN model result of \S\ref{subsec:basemodel} as the baseline.

%From Table \ref{tab:base}, the base DNN model can predict the low-risk and stroke state well, while the results of medium- and high-risks are slightly weaker. In addition, the accuracy of the four classification prediction  is 81.23\%. 
\begin{table}[!h]
	\caption[]
	{classification report of base DNN model and random forest. }
	\label{tab:base}
	\centering
	\begin{tabular}{|c|c|c|c|c|c|}
		\hline
		Model& Risk state & Precision(\%) & Recall(\%) & F1 score(\%) & Support \\
		\hline
		\multirow{4}{*}{Base DNN}&low & 81.90 & 92.06 & 86.68 & 1096  \\
		\cline{2-6}
		&medium & 80.77 & 78.21 & 79.47 & 881  \\
		\cline{2-6}
		&high & 79.57 & 72.32 &  75.77&813  \\
		\cline{2-6}
		&attack & 84.38 &\uline{\textbf{74.48 }}& 79.12 & 290  \\
		\hline
		\multirow{4}{*}{Random Forest}&low&84.47&96.26&89.98&1096\\
		\cline{2-6}
		&medium&86.88&87.97&87.42&881\\
		\cline{2-6}
		&high&86.13&79.46&82.66&813\\
		\cline{2-6}
    		&attack&91.53&\uline{\textbf{59.66}}&72.23&290\\
		\hline
	\end{tabular}
\end{table}

\subsubsection{Explanation of feature importance}
\label{sssec:qifeature_importance}

\begin{figure}[!h]
	\begin{center}
		\includegraphics[width=7cm]{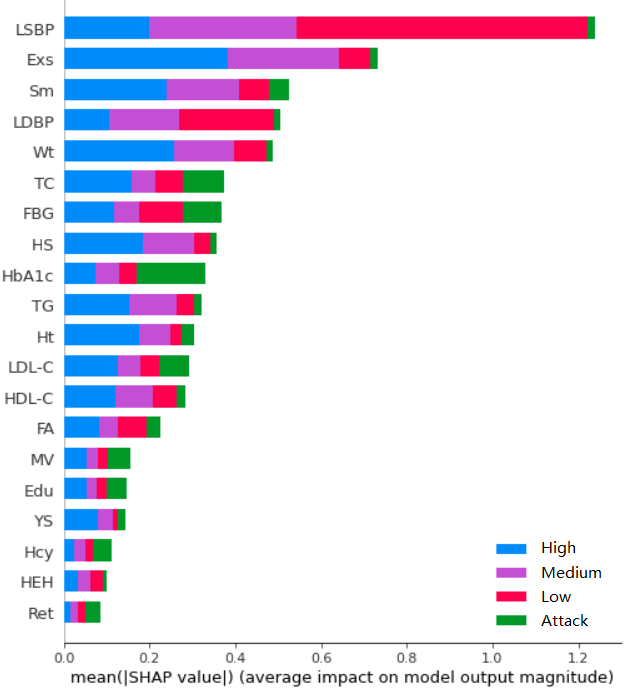}
		\caption{Ranking of feature importance. }
		\label{fig:RFC}
	\end{center}
\end{figure}

According to the average Shapley absolute values of every feature in the samples, the feature importance ranking of the base DNN model can be computed. Figure \ref{fig:RFC} is the feature importance diagram of the base DNN model. 
Different from \cite{series}, we provide not only the importance of features in stroke risk prediction, but also those for each state. For the four classification model, 
the features are listed with descent importance from top to bottom. 
Different colors in the graph represent the importance of features in different states of prediction.
Overall,   left systolic blood pressure (LSBP), lack of Exercise (Exs) and smoking (Sm) are the top3 features in stroke risk assessment. At the same time,  the important features for identifying different risk states could be also be found according to the SHAP values. 
 To name a few, LSBP, LDBP, FBG and Exs dominate the importance of low-risk prediction; LSBP, Exs, Sm and LDBP contributes more to the medium-risk prediction; Exs, Wt, Sm and LSBP play significant rules in determine the high-risk prediction; while for the prediction of attack state, HbA1c, TC, LDL-C has greater marginal effect. The explanation of selected features are shown in Appendix Table \ref{tab:feature exp}. 
%For example, LSBP is the most important feature in low-risk prediction, and its influence to the absolute shapley value of prediction for low-risk state could be measured by the value 0.55 (0.92-0.37) on average. While for the prediction of stroke state, HbA1c, TC, LDL-C has greater marginal effect.

\subsection{QIDNN: improvement of base DNN model}
\label{subsec:qidnn}

\subsubsection{Order-2 feature selection}
\label{sssec:interaction}

\begin{figure}[!p]
% 	\centering
% 	\subfigure[]{
% 		\label{fig:inter_all}
% 		\includegraphics[width=14cm]{fig/interaction_total2_ps.png}}
% 	\hspace{0.2cm} 
	\centering
	\subfigure[]{
		\label{fig:depend_c}
		\includegraphics[width=7cm]{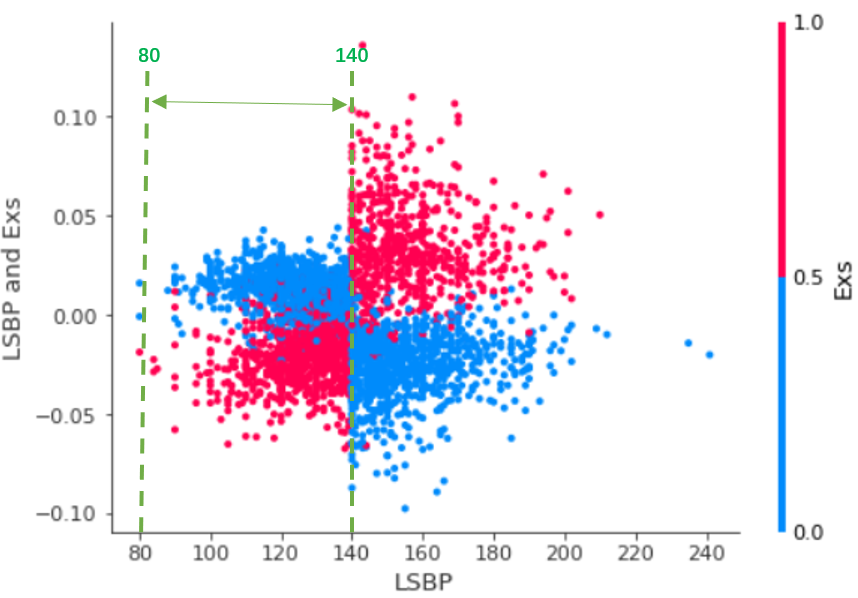}}
	\hspace{0.2cm} 
	\centering
	\subfigure[]{
		\label{fig:depend_d}
		\includegraphics[width=7cm]{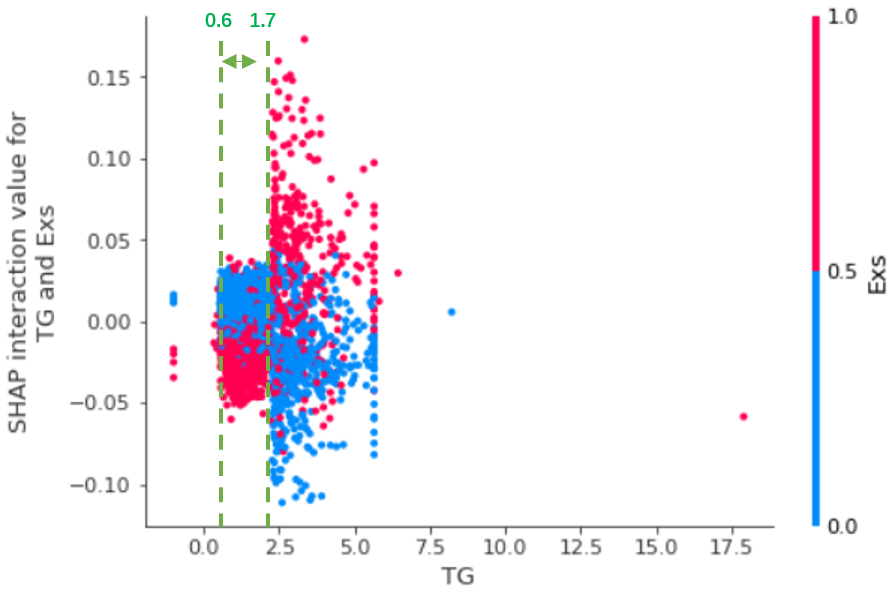}}
	\hspace{0.2cm} 
	\centering
	\subfigure[]{
		\label{fig:depend_e}
		\includegraphics[width=7cm]{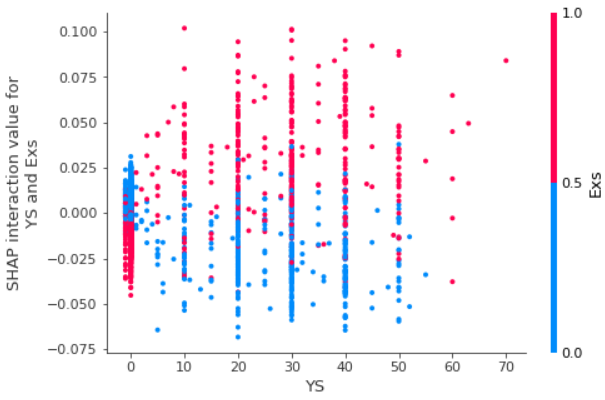}}
	\hspace{0.2cm} 
	\centering
	\subfigure[]{
		\label{fig:depend_g}
		\includegraphics[width=7cm]{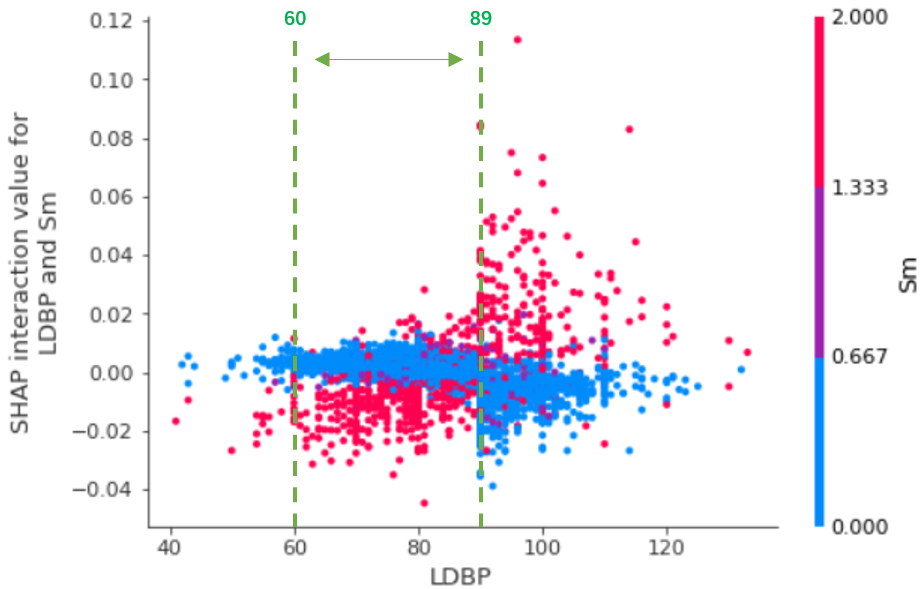}}
	\hspace{0.2cm} 
	\caption{ SHAP dependence calculations of the high risk explaination.   (a) SHAP denpendence between Exs and LSBP. (b) SHAP denpendence between Exs and TG. (c) SHAP denpendence between Exs and Ys. (d) SHAP denpendence between Sm and LSBP. The normal ranges are illustrated by the green dash lines. 
	}
	\label{figs:depend}
\end{figure}
It is suggested that there are some synergistic effects between different  risk factors \cite{pezzini2004synergistic}.  
So in order to improve the prediction accuracy,  the interactions between different features, i.e.  order-2 interaction values, are also taken as inputs of QI layer of QIDNN. The order-2 features are screened by the largest interaction and dependency value calculated by SHAP. Through considering both the results of feature importance and interaction strength of four states, the pairwise interactions between LSBP, Exs, Sm, LDBP, RSBP, HbA1c and HS are selected as the inputs of QI layer.

%It's necessary to consider the influence of the interaction between features which could enrich the doctors' advice in clinical diagnosis. By selecting the features with the largest interaction and dependency value calculated by SHAP, we can get the order-2 interaction value by inputting them to QI layer of QIDNN to improve the performance of the model. 
We use the high-risk state explainer as an example to show some results intuitively in Figure \ref{figs:depend}. 
% Figure \ref{figs:depend} depicts the interaction values and SHAP dependency for high-risk state prediction. In Figure \ref{fig:inter_all}, fast exact computation of pairwise interactions are implemented and showed. SHAP interaction values are a generalization of SHAP values to higher order interactions, where the main self-interaction effects are on the diagonal and the interaction effects are off-diagonal. Features are arranged in descending order of interaction value. It shows that Exs, BMI, TG and LSBP has large interaction strength in high-risk state prediction. Figure \ref{fig:depend_c}-\ref{fig:depend_g} are SHAP dependence plots showing how a feature's value (x-axis and y-axis on the right) impacts the prediction (y-axis on the left) of every sample (each dot) in a dataset.
Figure \ref{fig:depend_c}-\ref{fig:depend_e} describe the interactions between Exs and LSBP, TG, YS, respectively. The normal range of the indicators in Table \ref{tab:feature exp} are also illustrated in these figures (green dotted lines).  We can see that when the indicators, sorted along the x-axis are not in its normal range, lack of exercise (y = 1) will promote patients to be high-risk. Figure \ref{fig:depend_g} shows the interaction between LDBP and Sm. It tells that non-smoking has no effect on high-risk, while smoking increases the probability of high-risk, especially when the blood pressure is raising.
It is worth noting that, from Figure \ref{fig:depend_c}, when the indicators are normal, lack of exercise seems to hinder the patient from suffering the high-risk. In this case, the risk level is not reduced directly. It means under the current data,   the subject with normal blood pressure and lack of exercise  has less probability to be   high-risk state.

%Through the analysis of the interactions between features, combinative features can be added to the input layer to improve the preformance of the model. It can enrich the doctors' advices. For example, because there is an interactive relationship between blood pressure and smoking, if the blood pressure does not fall for a long time and the patient's risk level is high, it is strongly recommended that the patient should give up smoking.

\subsubsection{Numerical results of QIDNN}

%We can not compute order-2 interactions of all the features  since the limited samples can hardly ensure the convergence of the model with increasing parameters introduced by interactions computation.
 The improvement results are shown in Table \ref{tab:QIDNN}. For QIDNN model, the number of features which is showed as  ``34+N" means the features are composed of 34 original features and N order-2 interaction features. It can be seen from the table that with the increasing number of order-2 interaction features,
the mean of the recall increases. Specifically, we list the recall value of the attack state for each iteration. It confirms that QIDNN with the increasing number of interaction features could improve the recall of the attack state. What's more, the selected  order-2 features also accelerate the convergence, decrease  the loss of the model, and improve the overall accuracy. Hence, the performance of the model is improved by adding order-2 interactions of features. 
%In addition, after adding 7 combinative features, there is no more gain by adding more combinative features, since over-fitting may occur due to the increasing number of parameters. %Adding more features can not guarantee the convergence of the model.
\begin{table}[!h]
	\caption[]
	{Influence of the number of interaction features on performance of QIDNN. }
	\label{tab:QIDNN}
	\centering
	\begin{threeparttable}
	\begin{tabular}{|c|c|c|c|c|c|c|}
		\hline
		\tabincell{c}{Model} & \tabincell{c}{\#Features\tnote{1}} & 
		\tabincell{c}{Mean(R)(\%)}\tnote{2} &
		\tabincell{c}{R(s)(\%)}\tnote{3} &
		\tabincell{c}{Accuracy(\%)} &
		\tabincell{c}{\#Iterations\tnote{4}} & \tabincell{c}{Min(loss)\tnote{5}} \\
		\hline
		DNN & 34 &79.27 & 74.48 & 81.23 & 105 & 0.6122  \\
		\hline
		\multirow{5}{*}{QIDNN} & 34+3  & 79.46 & 74.83& 81.59& 55 & 0.5243  \\
		\cline{2-7}
		 & 34+4  & 79.85 & 74.83 & 82.01 & 34 & 0.5241  \\
		\cline{2-7}
		 & 34+5  & 80.82 & 76.55 & 82.37 & 24 & 0.5145  \\
		\cline{2-7}
		 & 34+6  &80.85 &77.24 & 82.60 & 15 & 0.5135  \\
		\cline{2-7}
		 & 34+7  & 81.60&78.28 & 83.25 & 17 & 0.5038  \\
		\hline
	\end{tabular}
	\begin{tablenotes}
	\footnotesize
		\item[1] \#Features: number of features.
		\item[2] Mean(R): mean of the recalls on all four states.
		\item[3] R(s): recall of the attack state.
		\item[4] \#Iteration: number of iterations.
		\item[5] Min(loss): minimum of loss value.
	\end{tablenotes}
	\end{threeparttable}
\end{table}

% \subsubsection{Explanation of state and state transition}
% \label{sssec:Exp_state}
% \begin{figure}[!bp]
% 	\centering
% 	\subfigure[]{
% 		\label{fig:lh_l}
% 		\includegraphics[width=14cm]{fig/single_lh_low.png}}
% % 	\hspace{0.2cm} 
% % 	\centering
% % 	\subfigure[]{
% % 		\label{fig:lh_m}
% % 		\includegraphics[width=14cm]{fig/single_lh_med.png}}
% 	\hspace{0.2cm} 
% 	\centering
% 	\subfigure[]{
% 		\label{fig:lh_h}
% 		\includegraphics[width=14cm]{fig/single_lh_high.png}}
% % 	\hspace{0.2cm} 
% % 	\centering
% % 	\subfigure[]{
% % 		\label{fig:lh_s}
% % 		\includegraphics[width=14cm]{fig/single_lh_s.png}}
% %	\hspace{0.2cm} 
% 	\caption{Explanation of stroke risk prediction on a sample of low-risk. (a) Force plot on model explainer of low risk.
% 	%(b) Force plot on model explainer of medium risk.
% 	(b) Force plot on model explainer of high risk. 
% 	%(d) Force plot on model explainer of attack state.
% 	}
% 	\label{figs:lh}
% \end{figure}
 
\begin{figure}[!h]
	\begin{center}
		\includegraphics[width=14cm]{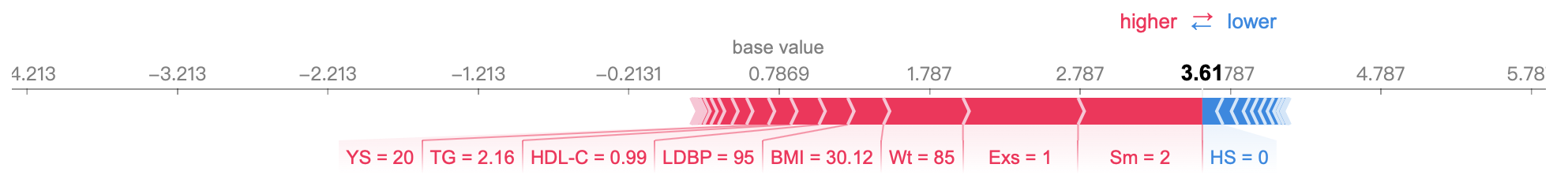}
		\caption{Explanation of stroke risk prediction on a sample of high-risk. }
		\label{fig:h}
	\end{center}
\end{figure}

\begin{figure}[!tp]
	\centering
	\subfigure[]{
		\label{fig:hs_h}
		\includegraphics[width=14cm]{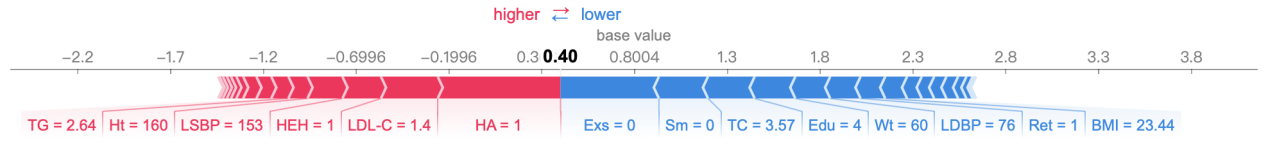}}
	\hspace{0.2cm} 
	\centering
	\subfigure[]{
		\label{fig:hs_s}
		\includegraphics[width=14cm]{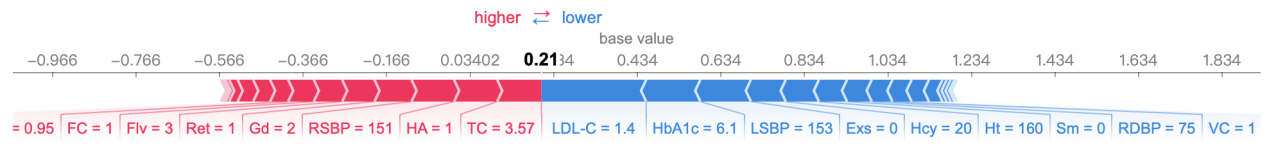}}
	\hspace{0.2cm} 
	\caption{Explanation of stroke risk prediction on a sample of high-risk. (a) Force plot on model explainer of high risk. (b) Force plot on model explainer of attack state.}
	\label{figs:hs}
\end{figure}

In order to increase the interpretability and practicability of the model, the SHAP DeepExplainer for the QIDNN model is established to identify the dominate risk factors leading to state transition for each subject. 
Here we use two examples to explain how it works.
 In the first example, we randomly select a sample of high-risk. The predicted value for the four states is (0,0,3.6051,0)\footnote{According to \eqref{eqn:yqidnn}, the predicted value of the four states should sum up to 1, however, in the numerical experiment, we use the value before {\it softmax}, due to the direct use of the CELoss of PyTorch to avoid vanishing gradient.},  which means that the predicted state being high-risk with the maximum score 3.6051, and no state transition potential.
Through the single sample analysis tool of SHAP, we obtain four explainers corresponding to the four states. 
These four explainers take a single sample as input, and provide the force graph of each state. Figure \ref{fig:h} shows the force graph obtained by explainers corresponding to  high-risk state. In the figure, Shapley values of feature attributes are visualized as ``forces", and each feature value is a force that increases or decreases the value of the prediction. The prediction starts from the base value, which is the average of the predictions, and each Shapley value is presented as an arrow to increase (red) or decrease (blue) the prediction.

Figure \ref{fig:h} describes that smoking behavior (even worse 20 years of smoking), lack of exercise, higher BMI, and the abnormal indicators, such as blood pressure, promotes this patient with current state of high-risk; while no family history of stroke hinders the patient from belonging to high-risk.  

Figure \ref{figs:hs} is an example of an analysis of high-risk patients. The predictive value for four states is (0,0,0.40,0.21), which means the patient having a tendency to turn from high-risk to attack. Combined with Figure \ref{fig:hs_h} and \ref{fig:hs_s}, it can be seen that the state is high-risk due to the family history of stroke and hypertension, a high blood pressure and older age at present (Ret=1). However, the patient does not smoke, and keeps exercising to ensure the normal BMI and TC, which reduce the risk. In particular, we can see that the patient has a higher education, better understanding and higher attention of the disease would probably reduce the risk of attack. This is consistent with the conclusion of the article \cite{liu2007stroke}, i.e. the prevention of stroke can be accomplished by better and earlier treatment of hypertension and health education. Combined with the analysis results, we suggest that this patient should pay attention to the light diet (Flv), and take drugs to ensure the normal blood pressure and cholesterol.

%SHAP DeepExplainer is estabilished based on trained model. The better the performance of prediction model, the more reasonable the explanation on current state and state transition. Therefore, we keep improving the performance of model prediction in \S\ref{subsec:qidnn} and \S\ref{subsec:mmoestroke}.
\subsection{MMOE: proceeding improvement of attack state}
\label{subsec:mmoestroke}
\subsubsection{Model selection for auxiliary objective}
Table \ref{tab:stroke} shows the prediction results of current popular algorithms for stroke occurrence prediction on our dataset. Apart from the indicators  mentioned above,  AUC, the area under the ROC curve, is taken into consideration to evaluate the performance of binary classifiers in the medical diagnosis domain \cite{ROC}.
AUC plays the role of indicating whether the sorting is correct i.e. whether the scores of positive samples are greater than the negative samples as it considers both sensitivity and specificity \cite{khosla2010integrated}, and it does not depend on the selection of thresholds.
%\textcolor{red}{ For other indicators, we choose 0.5 as the threshold, which means if the predicted score is greater than 0.5, the model regards the sample as stroke.} 
We can see from Table \ref{tab:stroke}, DNN-B (B is short for binary to differ from the base DNN model for four classification) is the best under the evaluation of all indicators, so it  is chosen as the Expert1 of MMOE. Compared with LR, Gradient Harmonizing Mechanism Logistic Regression (GHMLR) \cite{2018Gradient} outperforms on imbalance sample set of stroke occurrence (1:10 as we mentioned) by using gradient density as a hedging to disharmonies between different examples. While RF is greater than GHMLR as its precision, recall and f1 score are better with similar AUC values. Therefore AUC could not be treated as unique evaluating indicator in our project. 
 %Furthermore, compared with GHMLR, we do not need to adjust threshold by using RF method. This is also a property that a good binary classification model needs to meet.

\begin{table}[!h]
	\caption[]
	{Comparison on binary classifications of stroke occurrence prediction. }
	\label{tab:stroke}
	\centering
	\begin{tabular}{|c|c|c|c|c|}
		\hline
		Model & Precision(\%) & Recall(\%) & F1 score(\%) & Auc \\
		\hline
		LR & 62.03 & 64.49 & 63.23 & 0.8681  \\
		\hline
		GHMLR & 78.60 & 73.11 & 75.76 & 0.9395  \\
		\hline
		GBDT & 100 & 53.42 & 69.64 & 0.9057  \\
		\hline
		RF & 97.92 & 78.17 & 86.94 & 0.9370  \\
		\hline
		DNN-B & 93.37 & 82.49 & 87.59 & \uline{\textbf{0.9781}}  \\
		\hline
	\end{tabular}
\end{table}
\subsubsection{Numerical results of MMOE}
The comparison between MMOE and single objective realization  are shown in Table \ref{tab:MMOE2} and Table \ref{tab:comparison}, respectively. Here, attach occurrence prediction is objective 1 and stroke risk assessment is objective 2. We would like to emphasize that, we select the top 20 features from original 34 features as the input to ensure the convergence. It explains why the performance of DNN-B in  Table \ref{tab:MMOE2} is worse than that in  Table \ref{tab:stroke}. It can be seen from the tables that both objectives of MMOE are improved compared with single objective model. 
For attach occurrence prediction, recall is increased by 10.6\% (from 73.86\% to 81.72\%) and AUC is increased by 1.6\% (from $97.30\%$ to $98.92\%$) with the similar  precision. 
Consider the stroke risk prediction, as shown in Table \ref{tab:comparison}. The overall prediction accuracy is $84.45\%$, which is 4.8\% higher than that of QIDNN model with feature structure of 20 + 3 (80.58\%), and even 3.2\% higher than that of QIDNN model with feature structure of 34 + 3 (81.85\%). Table \ref{tab:MMOE4} extends the third and the fifth rows of Table \ref{tab:comparison}, and the details shows that the prediction performance (precision, recall and F1 score) of each state is improved.  We believe that when the data set is large enough, the two objective optimization of stroke occurrence and stroke risk prediction based on all features of MMOE framework would achieve better results.

\begin{figure}[!tp]
	\centering
	\subfigure[]{
		\label{fig:loss_all}
		\includegraphics[width=7cm]{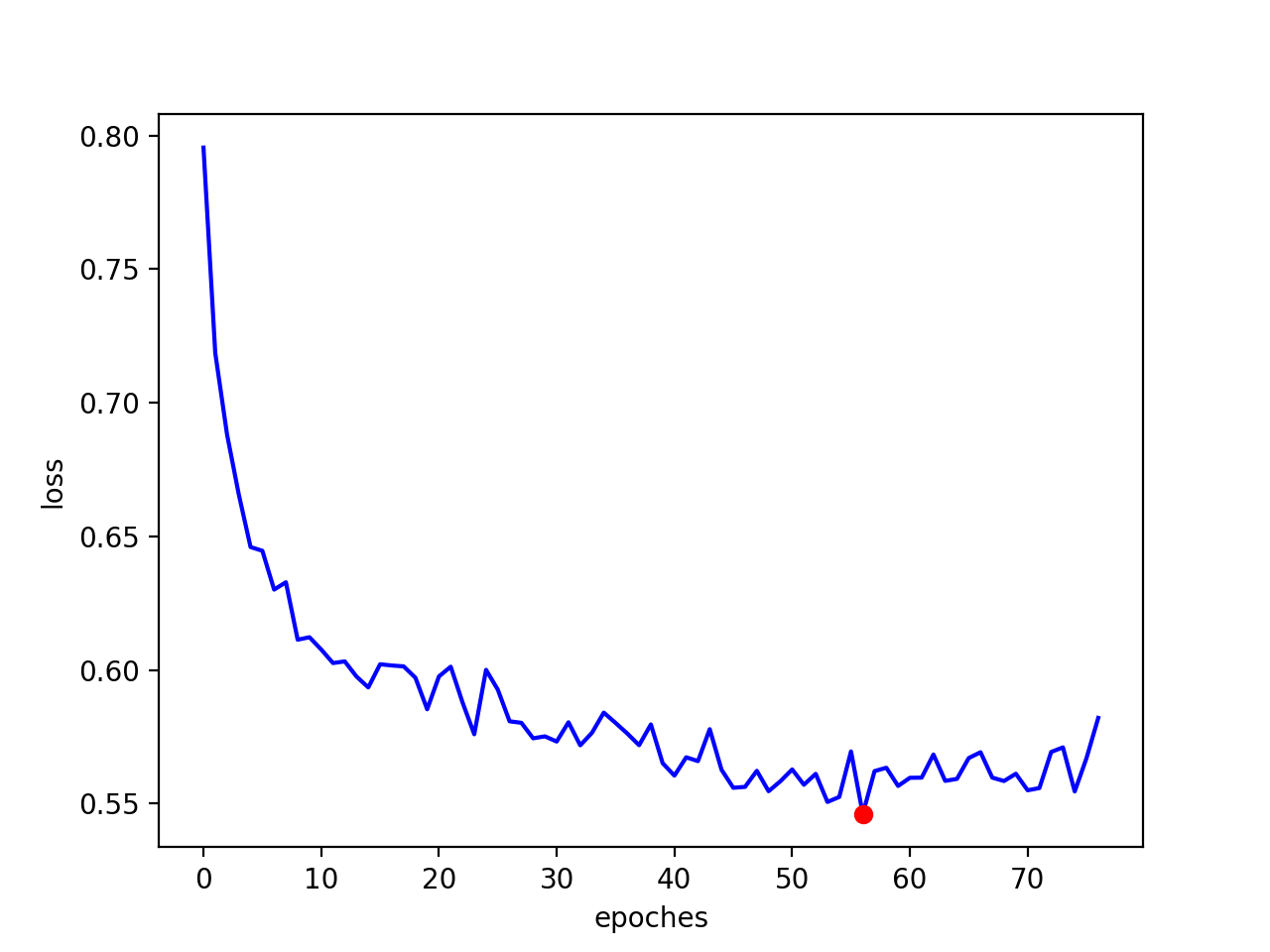}}
	\hspace{0.1cm} 
	\subfigure[]{
		\label{fig:auc_ep1}
		\includegraphics[width=7cm]{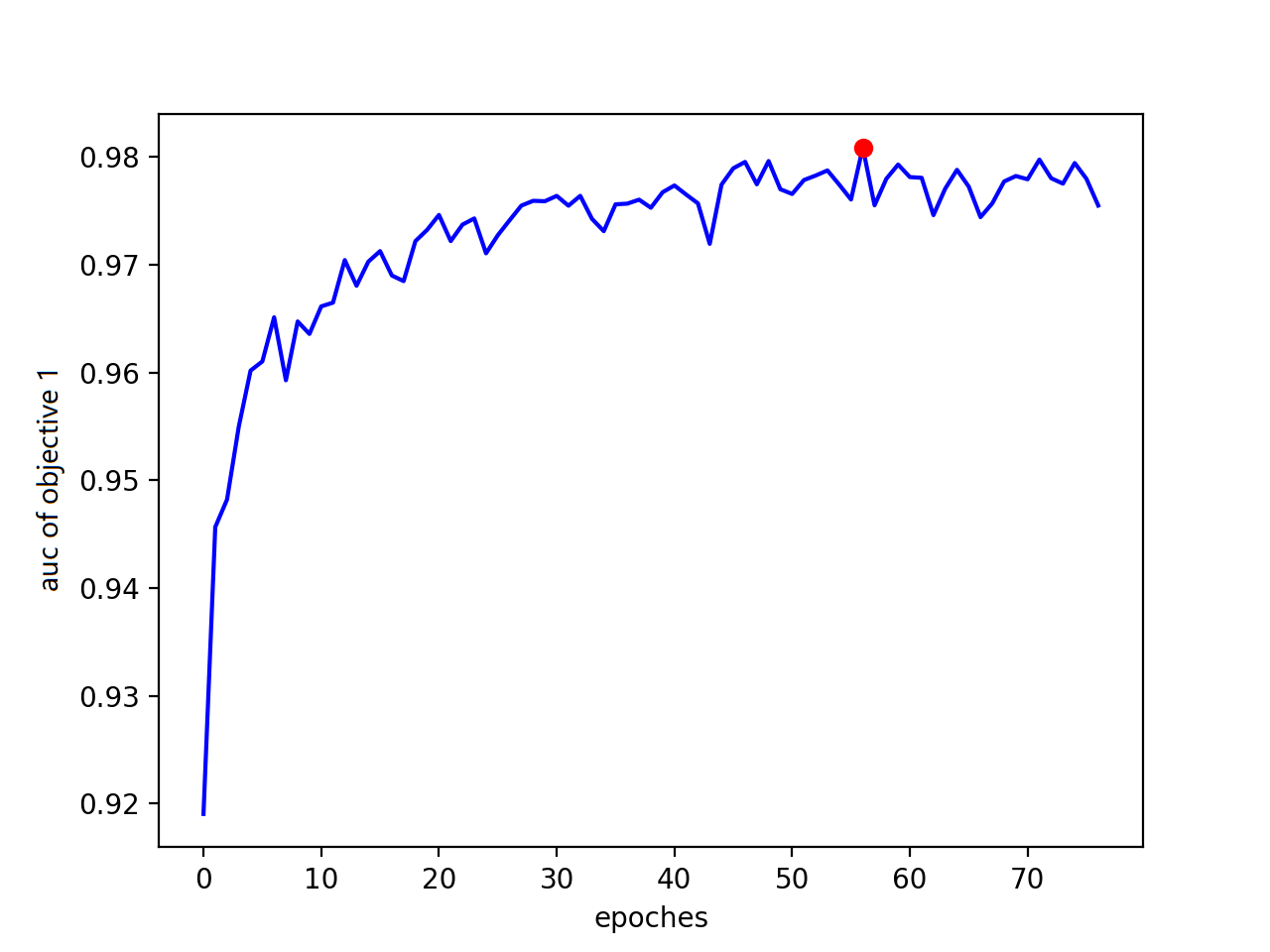}}
	\hspace{0.1cm} 
	\subfigure[]{
		\label{fig:loss_ep1}
		\includegraphics[width=7cm]{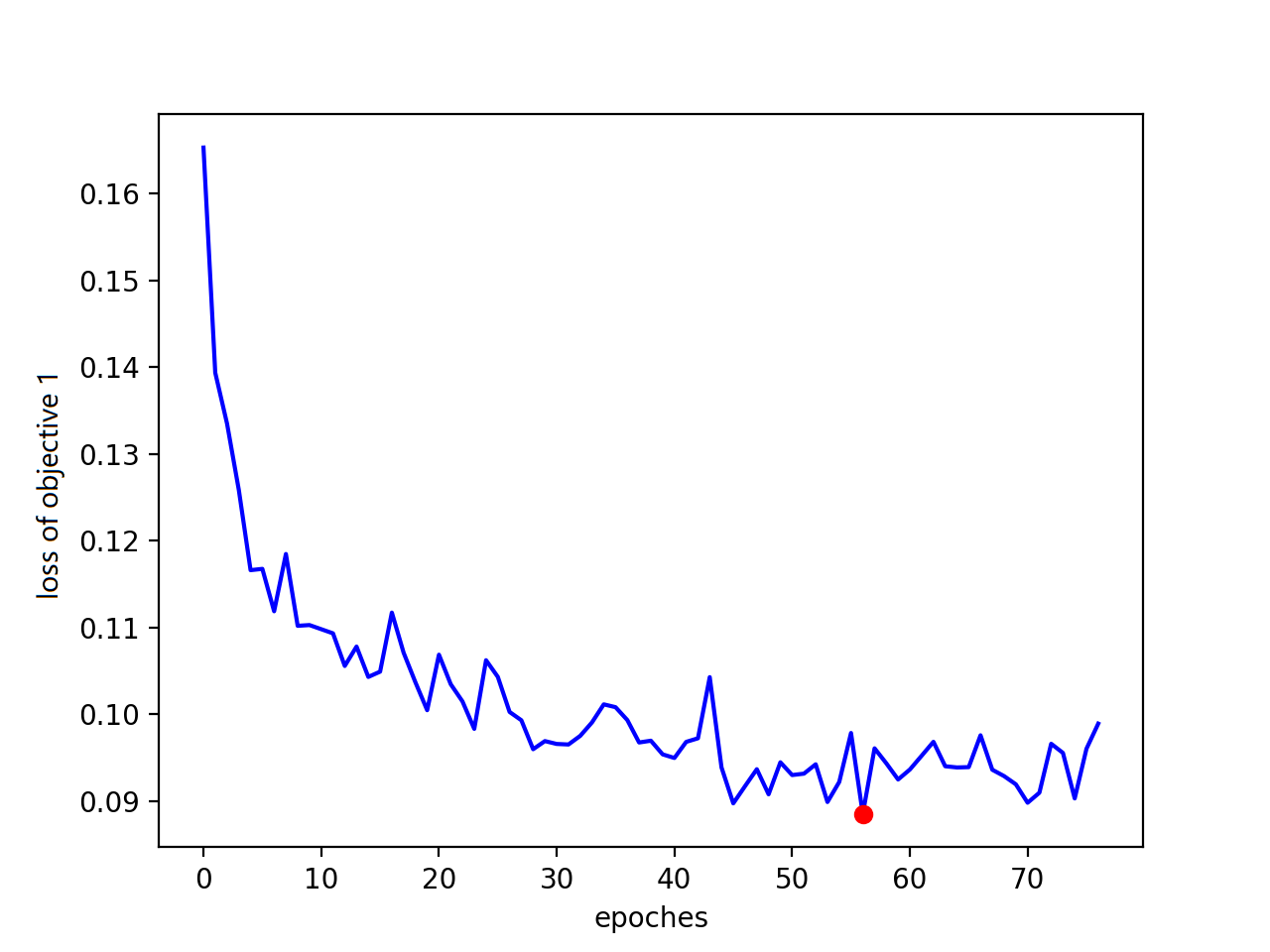}}
	\hspace{0.1cm} 
	\subfigure[]{
		\label{fig:loss_ep2}
		\includegraphics[width=7cm]{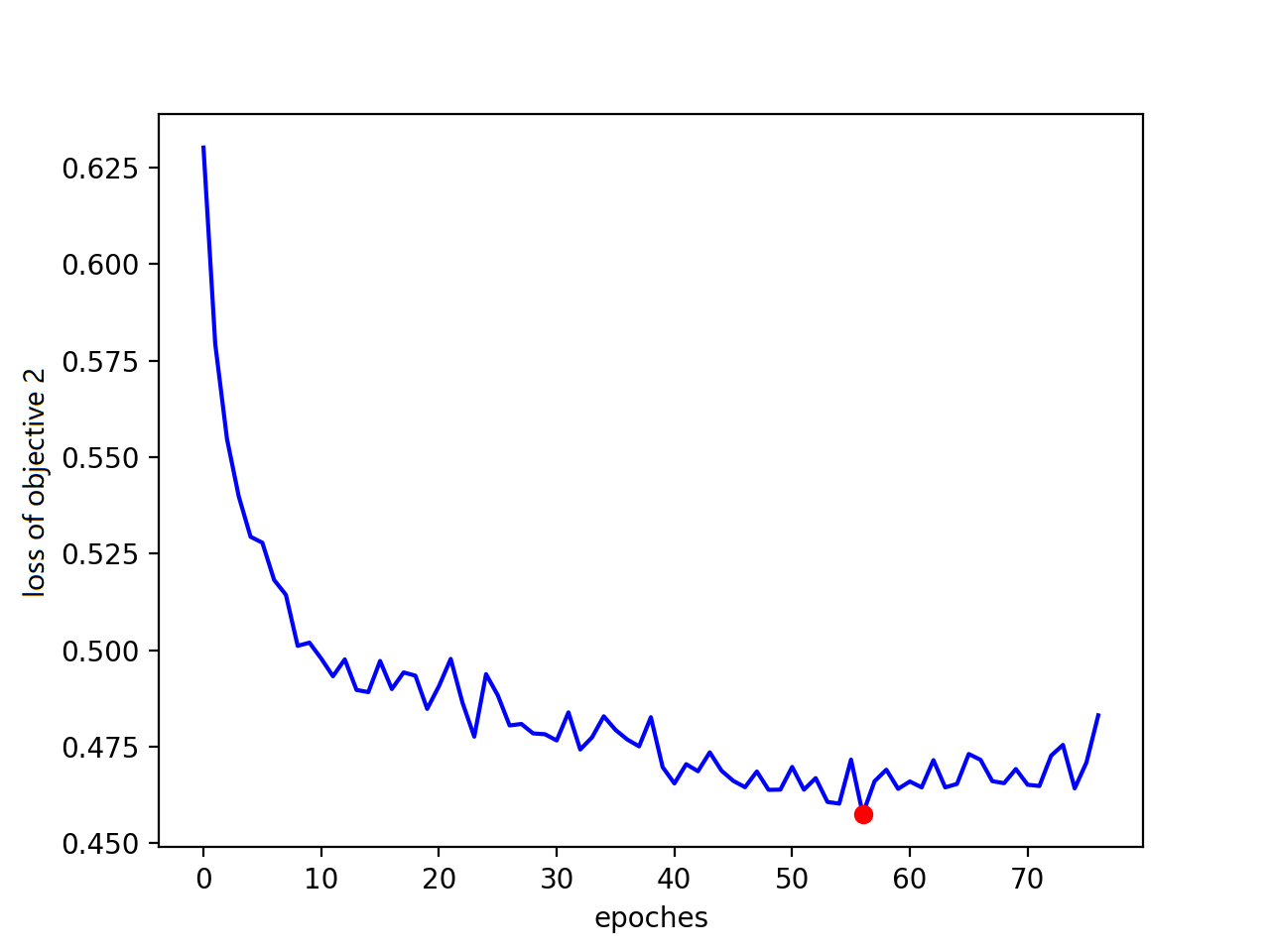}}
	\hspace{0.1cm} 
	\caption{Numerical results of MMOE. (a) The convergence of loss value for MMOE model. (b) AUC of the objective 1 of MMOE. (c) The convergence of loss value for objective 1 of MMOE. (d) The convergence of loss value for objective 2 of MMOE.}
	\label{figs:loss}
\end{figure}

Figure \ref{figs:loss} shows the total loss values, each objective's loss value and AUC of objective 1 of MMOE using top-20 features as input.
The red dot in the figure is the early stop point. It can be seen that under this model parameters, MMOE reaches the optimal after 73 epochs, the loss of each objective is convergent, and continuous training will lead to over-fitting.
\begin{table}[!h]
	\caption[]
	{Comparison between DNN and MMOE on stroke occurrence prediction. }
	\label{tab:MMOE2}
	\centering
	\begin{tabular}{|c|c|c|c|c|}
		\hline
		Model & Precision(\%) & Recall(\%) & F1 score(\%) & Auc \\
		\hline
		DNN-B & 92.75 & 73.86 & 82.23 & 0.9730  \\
		\hline
		MMOE &\uline{\textbf{ 92.94}} & \uline{\textbf{81.72}} &\uline{\textbf{86.97}} & \uline{\textbf{0.9892}}\\
		\hline
	\end{tabular}
\end{table}
\begin{table}[!h]
	\caption[]
	{Comparison among stroke risk prediction methods. }
	\label{tab:comparison}
	\centering
%	\begin{threeparttable}
	\begin{tabular}{|c|c|c|}
		\hline
		Model& \tabincell{c}{\#Features} & Accuracy(\%) \\
		\hline
		Base DNN & 20 & 79.51  \\
		\hline
		Base DNN & 34 & 81.59  \\
		\hline
		QIDNN & 20+3 & 80.58  \\
		\hline
		QIDNN & 34+3 & 81.85  \\
		\hline
		MMOE & 20+3& \uline{\textbf{84.45}}\\
		\hline
	\end{tabular}
%	\begin{tablenotes}
%	\footnotesize
%		\item[1] \#Features: number of features.
%	\end{tablenotes}
%	\end{threeparttable}
\end{table}

\begin{table}[!htpb]
	\caption[]
	{Comparison of MMOE with QIDNN on stroke risk prediction. }
	\label{tab:MMOE4}
	\centering
	\begin{tabular}{|c|c|c|c|c|c|}
		\hline
		Model & Risk state & Precision(\%) & Recall(\%) & F1 score(\%) & Support \\
		\hline
		\multirow{4}{*}{QIDNN}&low&82.64&93.80&87.86& 1096\\
		\cline{2-6}
		&medium&82.24&74.12&77.97&881\\
		\cline{2-6}
		&high&75.22&74.66&74.94&813\\
		\cline{2-6}
		&attack&83.83&67.93&75.05&290\\
		\hline
		\multirow{4}{*}{MMOE}&low & 84.13 & 94.34 & 88.95 & 1096  \\
		\cline{2-6}
		&medium & 83.60 & 81.61 & 82.60 & 881  \\
		\cline{2-6}
		&high & 84.31 & 74.05 &  78.85&813  \\
		\cline{2-6}
		&attack & 88.81 & 84.83 & 86.77 & 290  \\
		\hline
	\end{tabular}
\end{table}

On the premise of improving the overall effect of the model, we further compare the prediction results of each method for attack state, as shown in Table \ref{tab:MMOE3}. This is also the motivation of using the multi-objective model. In this case, the base DNN model  use 20 original features, QIDNN and MMOE use feature structure of 20 + 3. 
It can be seen from Table \ref{tab:MMOE3} that for the attack state, MMOE is much better than the single objective model (the base DNN model and the QIDNN model) with respect to all the indicators. 
%We choose MMOE as the final model since it can better predict the recall, which meets the needs of stroke diagnosis. Furthermore, four classification model could provide the patient's current risk state and state transition, and it can attribute the prediction results combined with SHAP.
For attack state, compared with QIDNN model, F1 score of MMOE is increased by 15.6\% (from 75.05\% to 86.77\%), the recall is increased by 24.9\% (from 67.95\% to 85.83\%).
\begin{table}[!h]
	\caption[]
	{Comparison of attack state prediction results. }
	\label{tab:MMOE3}
	\centering
	\begin{tabular}{|c|c|c|c|c|}
		\hline
		Model & Precision(\%) & Recall(\%) & F1 score(\%) \\
		\hline
		Base DNN & 86.51 & 64.14 & 73.66 \\
		\hline
		QIDNN & 83.83 & 67.93 & 75.05  \\
		\hline
	%	Expert1 of MMOE & 92.94 & 81.72 &  86.97  \\
	%	\hline
		MMOE & \uline{\textbf{88.81}} & \uline{\textbf{84.83 }}& \uline{\textbf{86.77}} \\
		\hline
	\end{tabular}
\end{table}

\section{Conclusion}
\label{sec:conc}
In view of the diagnosis and analysis of stroke,     stroke risk assessment models are proposed by deep learning method. Moreover, the proposed models could identify  the determinants of these risk states for every subject, which  makes agent-based treatment possible and   improve the effectiveness of stroke intervention and prevention.  
%For patients, accurate detection and early treatment can save expensive treatment costs. For doctors, the personalized attribution analysis of each patient risk state by determinators is more targeted and can be used as an experience supplement for disease diagnosis.

Specifically, for the accuracy of all stroke risk prediction, we propose a QIDNN model by adding quadratic interactive features into DNN to cope with small data set. This method can flexibly control the number of combination features, balance model parameters and the number of samples to ensure the model convergence. In addition, aiming at the problem that the method is not effective for the important attack state, this paper uses the multi-objective optimization framework of MMOE for reference, and builds a model with the objective of optimizing both stroke occurrence prediction and stroke risk prediction. In the case of the same feature, the prediction effect of each risk state was improved and the accuracy of the stroke risk assessment was improved by 4.8\% (from 80.58\% to 84.45\%), F1 score of attack state was increased by 15.6\% (from 75.05\% to 86.77\%) and the recall was increased by 24.9\% (from 67.95\% to 84.83\%) compared with QIDNN of single objective method. 
Finally, we note that this method can be applied to clinical prediction of other diseases, where missing data are common and risk factors are not well understood.
%\end{conclusion}
\maketitle

\section{Acknowledgement}
We thank Professor Tao Tang of United International College (UIC), for his suggestions on the manuscript. We also thank Professor Ivan Yuhui Deng and Professor Ping He from UIC for the helpful discussions.
\maketitle

%\bibliographystyle{plain}
%\bibliography{nonWave.bib}
\bibliographystyle{plain}
\bibliography{reference.bib}

\begin{thebibliography}{10}

\bibitem{series}
Stroke data analysis.

\bibitem{ahmad2018interpretable}
M.~A. Ahmad, A.~Teredesai, and C.~Eckert.
\newblock Interpretable machine learning in healthcare.
\newblock {\em 2018 IEEE International Conference on Healthcare Informatics
  (ICHI)}, page 447, 2018.

\bibitem{2017FSRP}
C.~Dufouil, A.~Beiser, L.A. McLure, and et~al.
\newblock Revised framingham stroke risk profile to reflect temporal trends.
\newblock {\em Circulation}, 135(12):1145--1159, 2017.

\bibitem{fatima2017survey}
M.~Fatima, M.~Pasha, and et~al.
\newblock Survey of machine learning algorithms for disease diagnostic.
\newblock {\em Int. J. Intell. Syst.}, 9(01):1, 2017.

\bibitem{ROC}
T.~Fawcett.
\newblock An introduction to roc analysis.
\newblock {\em Pattern. Recogn. Lett.}, 27(8):861--874, 06 2006.

\bibitem{2017DeepFM}
H.~Guo, R.~Tang, Y.~Ye, Z.~Li, and X.~He.
\newblock Deepfm: A factorization-machine based neural network for ctr
  prediction.
\newblock {\em Proceedings of the 26th International Joint Conference on
  Artificial Intelligence}, pages 1725--1731, 2017.

\bibitem{2019Machine}
J.~N. Heo, J.~G. Yoon, H.~Park, and et~al.
\newblock Machine learning–based model for prediction of outcomes in acute
  stroke.
\newblock {\em Stroke}, 2019.

\bibitem{2013QStroke}
J.~Hippisley-Cox, C.~Coupland, and P.~Brindle.
\newblock Derivation and validation of qstroke score for predicting risk of
  ischaemic stroke in primary care and comparison with other risk scores: A
  prospective open cohort study.
\newblock {\em Brit. Med. J.}, 346:f2573, 2013.

\bibitem{hung2017comparing}
C.~Y. Hung, W.~C Chen, P.~T. Lai, and et~al.
\newblock Comparing deep neural network and other machine learning algorithms
  for stroke prediction in a large-scale population-based electronic medical
  claims database.
\newblock In {\em 2017 39th Annual International Conference of the IEEE
  Engineering in Medicine and Biology Society (EMBC)}, pages 3110--3113. IEEE,
  2017.

\bibitem{khosla2010integrated}
A.~Khosla, Y.~Cao, C.~C. Lin, and et~al.
\newblock An integrated machine learning approach to stroke prediction.
\newblock {\em Proceedings of the 16th ACM SIGKDD international conference on
  Knowledge discovery and data mining}, pages 183--192, 2010.

\bibitem{2018Gradient}
B.~Li, Y.~Liu, and X.~Wang.
\newblock Gradient harmonized single-stage detector.
\newblock {\em ArXiv preprint}, 2018.

\bibitem{2011Stroke}
L.~Liu, D.~Wang, K.~Wong, and Y.~Wang.
\newblock Stroke and stroke care in china.
\newblock {\em Stroke}, 42(12):3651--3654, 2011.

\bibitem{liu2007stroke}
M.~Liu, B.~Wu, W.~Wang, L.~Lee, S.~Zhang, and L.~Kong.
\newblock Stroke in china: epidemiology, prevention, and management strategies.
\newblock {\em Lancet Neurol.}, 6(5):456--464, 2007.

\bibitem{lumley2002stroke}
T.~Lumley, R.~A. Kronmal, M.~Cushman, T.~A. Manolio, and S.~Goldstein.
\newblock A stroke prediction score in the elderly: validation and web-based
  application.
\newblock {\em J. Clin. Epidemio.}, 55(2):129--136, 2002.

\bibitem{NIPS2017}
S.~M. Lundberg and S.~Lee.
\newblock A unified approach to interpreting model predictions.
\newblock In I.~Guyon, U.~V. Luxburg, S.~Bengio, H.~Wallach, R.~Fergus,
  S.~Vishwanathan, and R.~Garnett, editors, {\em Advances in Neural Information
  Processing Systems 30}, pages 4765--4774. Curran Associates, Inc., 2017.

\bibitem{2017Consistent}
S.~M. Lundberg and S.~I. Lee.
\newblock Consistent feature attribution for tree ensembles.
\newblock {\em ArXiv preprint}, page 1706.06060, 2017.

\bibitem{2018MMOE}
J.~Ma, Z.~Zhe, X.~Yi, and et~al.
\newblock Modeling task relationships in multi-task learning with multi-gate
  mixture-of-experts.
\newblock {\em Proceedings of the 24th ACM SIGKDD International Conference on
  Knowledge Discovery and Data Mining}, pages 1930--1939, 2018.

\bibitem{2019largeEHR}
Mohammad, Taghi, Samadi, and et~al.
\newblock Development of an intelligent decision support system for ischemic
  stroke risk assessment in a population-based electronic health record
  database.
\newblock {\em Ecotoxicology and Environmental Safety}, 2019.

\bibitem{pezzini2004synergistic}
Alessandro Pezzini, Mario Grassi, Elisabetta Del~Zotto, Elena Bazzoli, Silvana
  Archetti, Deodato Assanelli, Nabil~Maalikjy Akkawi, Alberto Albertini, and
  Alessandro Padovani.
\newblock Synergistic effect of apolipoprotein e polymorphisms and cigarette
  smoking on risk of ischemic stroke in young adults.
\newblock {\em Stroke}, 35(2):438--442, 2004.

\bibitem{fm2010}
S.~Rendle.
\newblock Factorization machines.
\newblock {\em 2010 IEEE International Conference on Data Mining}, pages
  995--1000, 2010.

\bibitem{2017overviewMOOP}
S.~Ruder.
\newblock An overview of multi-task learning in deep neural networks.
\newblock {\em ArXiv preprint}, page 1706.05098, 2017.

\bibitem{shap1953definition}
L.~S. Shapley.
\newblock {\em A value for n-person games}, volume~28.
\newblock Princeton University Press, 1953.

\bibitem{2019survey}
Y.~Wang, Z.~Li, H.~Gu, and et~al.
\newblock China stroke statistics 2019: A report from the national center for
  healthcare quality management in neurological diseases, china national
  clinical research center for neurological diseases, the chinese stroke
  association, national center for chronic and non-communicable disease control
  and prevention, chinese center for disease control and prevention and
  institute for global neuroscience and stroke collaborations.
\newblock {\em Stroke Vasc. Neurol.}, 5(3):211--239, 2020.

\bibitem{1991FSRP}
P.~A. Wolf, R.~B. D'Agostino, Belanger~A. J., and Kannel~W. B.
\newblock Probability of stroke: A risk profile from the framingham study.
\newblock {\em Stroke}, 3(22):312--318, 1991.

\bibitem{2019Stroke}
X.~Xing, X.~Yang, F.~Liu, and et~al.
\newblock Predicting 10-year and lifetime stroke risk in chinese population the
  china-par project.
\newblock {\em Stroke}, 50(9):2371--2378, 2019.

\bibitem{2020RealTime}
J.~Yu, S.~Kwon, C.~Ho, C.~Pyo, and H.~Lee.
\newblock Ai-based stroke disease prediction system using real-time
  electromyography signals.
\newblock {\em Appl. Sci.}, 10(19):6791, 2020.

\bibitem{zhou2019mortality}
M.~Zhou, H.~Wang, X.~Zeng, and et~al.
\newblock Mortality, morbidity, and risk factors in china and its provinces,
  1990--2017: a systematic analysis for the global burden of disease study
  2017.
\newblock {\em Lancet}, 394(10204):1145--1158, 2019.

\end{thebibliography}

\newpage
\appendix
\section{Explanation of selected  features}
\begin{table}[!h]
	\begin{threeparttable}
	\caption[]
	{Explanation of selected features. }
	\label{tab:feature exp}
	\centering
	\begin{tabular}{|c|c|c|}
		\hline
		abbreviation &feature name & explaination \\
		\hline
		Arhm & arrhythmia  & 1: YES. 0: NO \\
		\hline
		BMI & Body Mass Index & normal range: 20-25\\
		\hline
		\tabincell{c}{Edu} & \tabincell{c}{Education}&
		\tabincell{l}{1: primary school and below.\\
			2: junior high school.\\
			3: technical secondary school\\ / senior high school.\\
			4: junior college / undergraduate.\\
			5: master degree or above.}\\
		\hline
		Exs & lack of Exercise & 1: YES. 0: NO\\
		\hline
		FA& Filling Age & -\\
		\hline
		FBG & Fasting Blood Glucose & normal range: 3.9-6.1mmol/L\\
		\hline
		HbA1c& glycosylated hemoglobin &normal range: 4\%-6\%\\
		\hline
		Hcy& Homocysteine & normal range: 5-15$\mathrm{\mu mol/L}$\\
		\hline
		\tabincell{c}{HDL-C} & \tabincell{c}{High Density Lipoprotein\\ Cholesterol} & \tabincell{l}{normal range for female:\\1.29-1.55 mmol/L;\\normal range for male:\\1.16-1.42 mmol/L}\\
		\hline
		HEH& History of Hypertension&1: YES. 0: NO\\
		\hline
		HS & History of Stroke & 1: YES. 0: NO\\
		\hline
		Ht/Wt & Height/ Weight & -\\
		\hline
		LDBP& Left Diastolic Blood Pressure & normal range: 60-89mmHg\\
		\hline
		\tabincell{c}{LDL-C}& \tabincell{c}{Low Density Lipoprotein\\ Cholesterol} &  \tabincell{c}{normal range: $<3.37mmol$}\\
		\hline
		LSBP& Left Systolic Blood Pressure & normal range: 80-140mmHg\tnote{1}\\
		\hline
		\tabincell{c}{MV}& \tabincell{c}{Meat and Vegetable}&\tabincell{l}{1: equilibrium.\\2: partial vegetable.\\3 partial meat.}\\
		\hline
		Ret & Retire&1: YES. 0: NO\\
		\hline
		Sm & Smoking &2: YES. 1: Quitting. 0: NO\\
		\hline
		TC&Total Cholesterol&3-5.2mmol/L\\
		\hline
		TG & triglyceride & normal range: 0.6-1.7mmol/L\\
		 
		\hline
		Ys & years of smoking & null for nonsmokers\\
		\hline
		
	\end{tabular}
	\begin{tablenotes}
	\footnotesize
	\item[1] The normal range of LSBP is 80-140mmHg and the optimal range is 80-120mmHg.
	\end{tablenotes}
		\end{threeparttable}
\end{table}

\end{document}